\documentclass[10pt,twocolumn,letterpaper]{article}

\usepackage{cvpr}
\usepackage{times}
\usepackage{epsfig}
\usepackage{graphicx}
\usepackage{amsmath}
\usepackage{amssymb}
\usepackage{multirow}

\begin{document}


\title{A Reference-Based 3D Semantic-Aware Framework for Accurate Local Facial Attribute Editing}

\author{Yu-Kai Huang, Yutong Zheng, Yen-Shuo Su, Anudeepsekhar Bolimera, Han Zhang, Fangyi Chen \\
and Marios Savvides\\
Carnegie Mellon University\\
5000 Forbes Ave, Pittsburgh, PA 15213\\
\tt\small yukaih2@andrew.cmu.edu, yutongzh@alumni.cmu.edu, ericsuece@cmu.edu,\\
\tt\small\{abolimer,hanz3,fangyic,marioss\}@andrew.cmu.edu
}

\twocolumn[{
\maketitle
\vspace{-0.53in}
\begin{center}
    \centering
    \includegraphics[width=1.42\columnwidth]{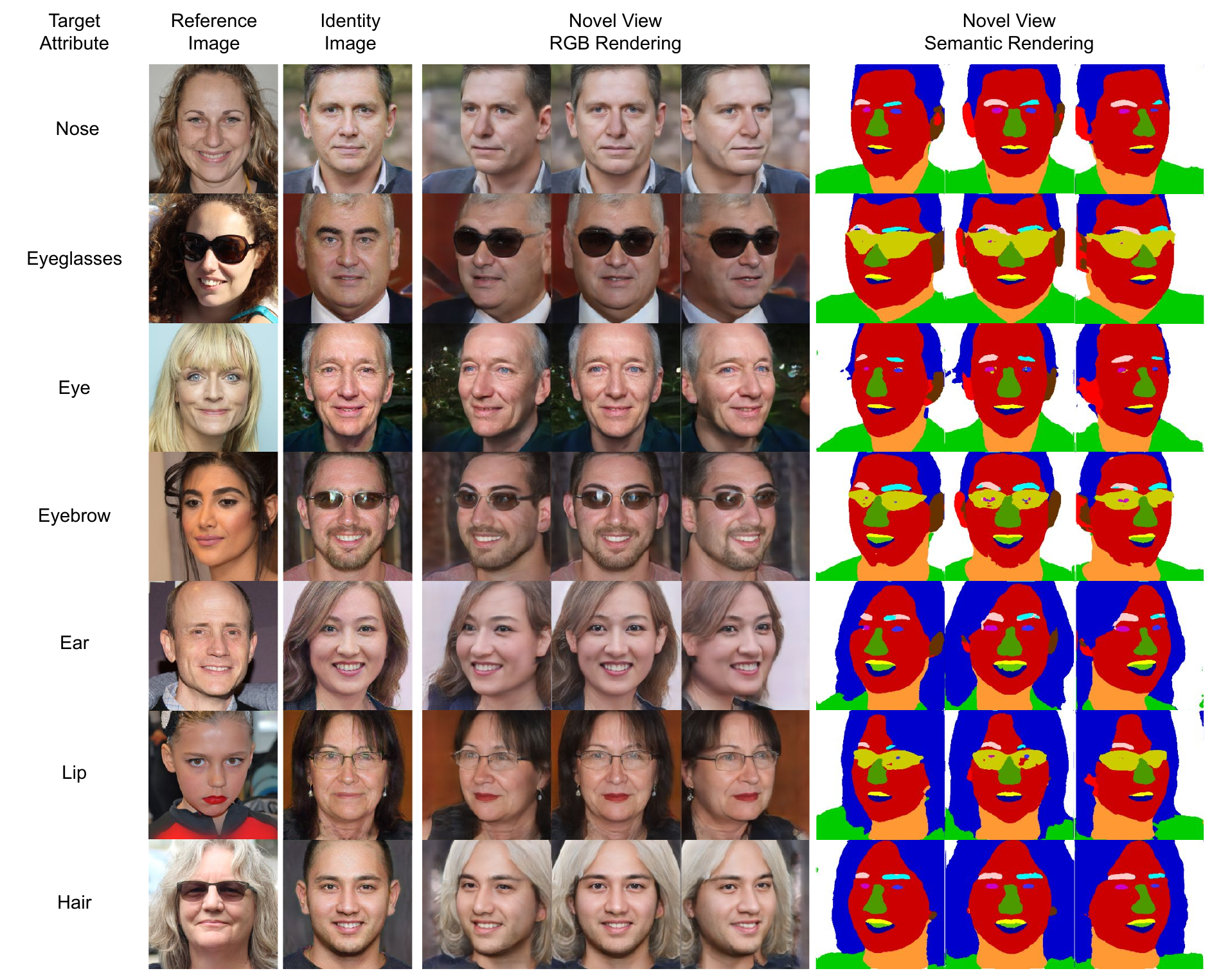}
    \captionof{figure}{Through our editing framework, we demonstrate seven types of local facial attribute edits, featuring novel-view RGB and semantic renderings. Our method consistently aligns semantic regions from the reference images with the identity image, ensuring accurate edits despite pose variations.}
    \label{fig:semantic_editing}
\end{center}
}]


\begin{abstract}

Facial attribute editing plays a crucial role in synthesizing realistic faces with specific characteristics while maintaining realistic appearances. Despite advancements, challenges persist in achieving precise, 3D-aware attribute modifications, which are crucial for consistent and accurate representations of faces from different angles. Current methods struggle with semantic entanglement and lack effective guidance for incorporating attributes while maintaining image integrity. To address these issues, we introduce a novel framework that merges the strengths of latent-based and reference-based editing methods. Our approach employs a 3D GAN inversion technique to embed attributes from the reference image into a tri-plane space, ensuring 3D consistency and realistic viewing from multiple perspectives. We utilize blending techniques and predicted semantic masks to locate precise edit regions, merging them with the contextual guidance from the reference image. A coarse-to-fine inpainting strategy is then applied to preserve the integrity of untargeted areas, significantly enhancing realism. Our evaluations demonstrate superior performance across diverse editing tasks, validating our framework's effectiveness in realistic and applicable facial attribute editing.

\end{abstract}

\vspace{-0.1in}
\section{Introduction}
\label{sec:intro}

Face synthesis~\cite{goodfellow2020generative,karras2019style,chan2022efficient,karras2017progressive,brock2018large,shen2018faceid,sun2023next3d} involves a range of techniques ranging from the creation of completely new faces that do not correspond to real individuals to the manipulation of existing facial attributes, such as changing nose, eyeglasses, hairstyle, etc. This field has seen significant developments, yet challenges remain, particularly in facial attribute editing. This process~\cite{shen2020interfacegan,simsar2023latentswap3d,patashnik2021styleclip,shen2021closed,zheng2021unsupervised,li2021image,dalva2022vecgan,bilecen2024reference} enhances control over changes, expressing diverse personalities and fashion styles while maintaining realistic appearances. Despite these advancements, achieving precise modifications of specific local facial attributes from a reference image in a 3D-aware manner remains a significant challenge. Such capability is crucial for applications in facial avatars and animations, where accurate 3D modeling is essential to view the face from multiple perspectives.

Building on existing methods, latent-based image editing methods~\cite{shen2020interpreting,patashnik2021styleclip} utilize GAN inversion technique~\cite{abdal2019image2stylegan,zhu2020domain,richardson2021encoding,wang2022high,alaluf2021restyle} to map an input image into the latent space ($W$, $W+$) of a generative model. This approach allows for the manipulation of the latent code to modify specific attributes of the image. However, identifying a latent code that accurately represents a local facial attribute from a reference image can be challenging due to semantic entanglement within the latent space~\cite{wei2023hairclipv2}. Additionally, these methods often lack the necessary contour and context guidance to effectively incorporate desired attributes from a reference while maintaining the overall structure and other attributes of the image. On the other hand, reference-based image editing methods~\cite{li2021image,dalva2022vecgan} focus on transferring specific attributes like texture, color, style, or structural elements from one image to another. Although powerful, these methods can lead to issues such as 3D inconsistencies, resulting in undesirable artifacts when images are viewed from different angles, primarily due to their lack of 3D-aware design~\cite{bilecen2024reference}.

In this paper, we introduce a novel framework designed to enhance reference-based, accurate local facial attribute editing that is both semantically precise and 3D-aware. Combining the strengths of latent-based and reference-based methods, our framework offers improved fidelity in attribute manipulation and ensures 3D consistency for multi-perspective viewing. The fundamental idea is the embedding of specific facial attributes from reference images into a tri-plane representation, utilizing a tri-plane encoding-decoding process. We blend these attributes with the identity's tri-plane representation to create modified tri-plane features for novel-view RGB and semantic image rendering. Our technique uses single-view reference images with predicted semantic masks to define the editing area and context while maintaining 3D consistency through feature blending in the tri-plane space. We then apply an inpainting technique to seamlessly integrate the reference context into the identity image, preserving the unaltered areas to produce the final output. Finally, we perform novel view rendering with these modified tri-plane features to generate both RGB and semantic images. This approach addresses previous challenges such as distortion, disruption of implicit 3D geometry after editing, inaccurate latent representations, and semantic entanglement in the latent space. Furthermore, our segmentation-aware design enables precise local editing, leveraging detailed information from reference images to guide the editing process. Our contributions are summarized as follows:









\begin{itemize}
    \item We introduce a novel framework for editing local facial attributes on a face image. This framework leverages single-view reference images as guides for appearance, and subsequently render the edited image's novel-view RGB image and semantic masks in a 3D-aware manner.
    \item We achieve the editing process by utilizing predicted semantic masks for precise manipulation and incorporating a blending process with tri-plane representation to align semantic regions from the reference image with the identity image in a 3D-aware manner. Additionally, we implement a coarse-to-fine inpainting pipeline to enhance context consistency, particularly along the borders of the edited areas, to smooth out abrupt transitions. Ultimately, the refined image is transformed into tri-plane features, which broadens its application potential.
    \item We achieve state-of-the-art performance on two benchmark tasks~\cite{bilecen2024reference}, showcasing the results in terms of image quality and preservation of unedited regions.

\end{itemize}


\section{Related Work}
\label{sec:related}



\begin{figure*}[t]
    \begin{center}
    \centering
    \includegraphics[width=1.0\linewidth]{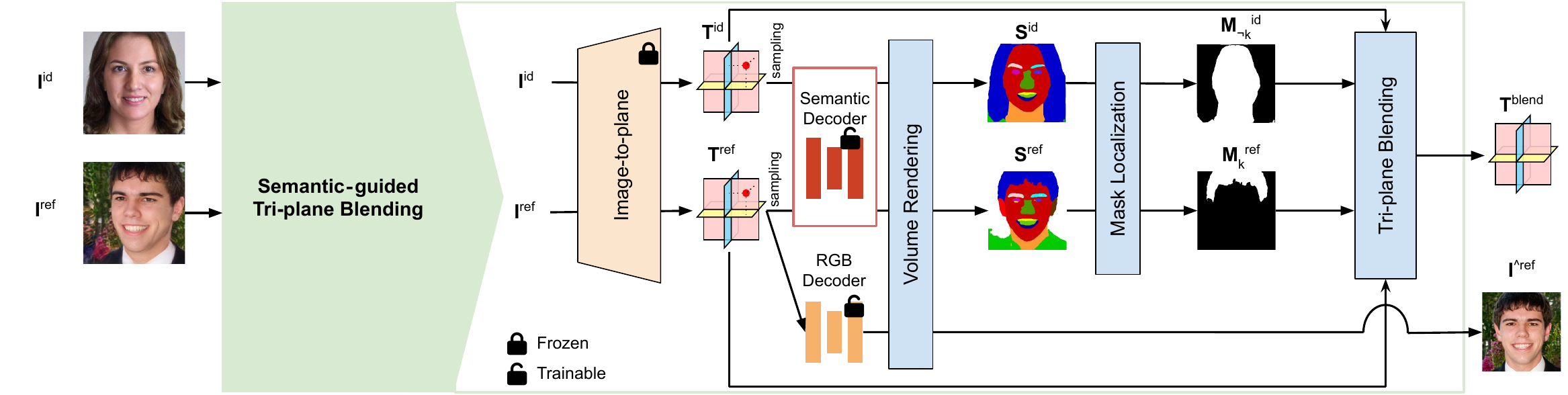}
    \end{center}
\vspace{-0.1in}
\caption{In the initial stage, we embed both the identity and reference images into the canonical tri-plane space, delineate the target editing area using predicted semantic masks, and incorporate features into the identity through alpha blending.}
\label{fig:pipeline_a}
\end{figure*}

\subsection{2D/3D GAN Inversion}
GAN inversion maps real images into the latent space of a pre-trained GAN to identify latent codes that can recreate these images, crucial for generating photorealistic faces with style control and realistic features. e4e~\cite{tov2021designing} explores this by examining StyleGAN's~\cite{karras2019style} latent space to enhance real image editing. However, traditional 2D GAN inversion struggles to capture the 3D structure of faces, important for rendering from new perspectives. In contrast, 3D GAN inversion~\cite{bhattarai2024triplanenet,yuan2023make,ye2024real3d} develops highly realistic 3D models by learning detailed object shapes, textures, and appearances. For instance, GOAE~\cite{yuan2023make} and TriPlaneNet~\cite{bhattarai2024triplanenet} employ an encoder-based approach with tri-plane representations from the EG3D family~\cite{chan2022efficient}. Real3D-Portrait~\cite{ye2024real3d} utilizes an image-to-triplane model with a motion adapter for enhanced reconstruction and animation capabilities. In our study, we use TriPlaneNet\cite{bhattarai2024triplanenet} as the 3D GAN inversion module, enabling direct 3D editing within the tri-plane space.

\subsection{Latent-based Image Editing}
Latent-based image editing leverages the latent representations within a generative model like StyleGAN~\cite{karras2019style} to enable precise and targeted modifications of image attributes by identifying interpretable directions in the latent space. Various techniques~\cite{harkonen2020ganspace, abdal2021styleflow, wu2021stylespace, richardson2021encoding, zheng2021unsupervised, wu2022hairmapper, shen2020interfacegan, patashnik2021styleclip,pehlivan2023styleres} have been crafted for StyleGAN to manipulate these directions. For example, InterFaceGAN~\cite{shen2020interfacegan} uses a conceptual hyperplane in the latent space to differentiate attributes, while StyleCLIP~\cite{patashnik2021styleclip} employs text-guided manipulation techniques such as latent optimization and mapping, greatly enhancing image editing possibilities beyond just annotated data. However, these methods encounter challenges, especially in accurately determining latent codes for specific facial attributes due to semantic entanglement in the latent space~\cite{wei2023hairclipv2, wu2022hairmapper}.
\subsection{Reference-based Image Editing}
Reference-based image editing~\cite{isola2017image,wang2018high,choi2020stargan,li2021image,dalva2022vecgan} uses reference images to guide the editing process for achieving specific visual effects. Techniques like HiSD~\cite{li2021image} and VecGAN++\cite{dalva2022vecgan} utilize image-to-image translation to facilitate editing, adding precise control over facial attributes through interpretable latent directions. However, these methods mainly handle 2D images and struggle with consistent 3D representations across multiple views. To overcome this, Bilecen \textit{et al.}~\cite{bilecen2024reference} introduced a reference-based, 3D-aware editing approach using the triplane latent spaces of ED3D, closely aligned with our methods. This approach enhances 3D consistency by encoding triplane features, facilitating spatial disentanglement, and automating feature localization within the triplane domain. Despite its advancements, their method relies on an external face segmentation network to identify attribute masks, requiring iterative processes for multi-view rendering in the image domain, which complicates optimization. In contrast, our method produces continuous semantic renderings across various poses from tri-plane space directly. We also implement a coarse-to-fine inpainting strategy that significantly boosts realism and image fidelity.

\section{Method}
\label{sec:method}

Our framework is outlined in three key stages: \textbf{Semantic-guided Tri-plane Blending}, \textbf{Coarse-to-fine Image Inpainting}, and \textbf{Interpretable Tri-plane Rendering}, which are detailed in Sec.~\ref{method:semantic_guided_triplane_blending},~\ref{method:coarse_to_fine_image_inpainting}, and~\ref{method:interpretable_triplane_rendering}, respectively.

\subsection{Semantic‐guided Tri-plane Blending}
\label{method:semantic_guided_triplane_blending}

In the process of combining facial attributes from a reference image $\mathbf{I}^{ref}$ into the identity image $\mathbf{I}^{id}$, the initial challenge is \textit{how to correctly align semantic regions from the reference image with the identity image}, especially when the identity and reference images do not share identical poses. A straightforward approach would be to crop and align the image using facial landmarks, centering the facial region within the image. However, this type of 2D alignment could lead to distortion and disrupt the implicit 3D geometry, making it unsuitable for rendering novel views in our application. Barbershop~\cite{zhu2021barbershop} introduces a two-stage alignment process that aligns each image by manipulating latent codes to reconstruct facial details, followed by aligning to match the target segmentation. Nonetheless, this approach may encounter problems with inaccurate latent representation and semantic entanglement mentioned in Sec.~\ref{sec:intro}.

\paragraph{Image-to-plane.} Image-to-plane~\cite{ye2024real3d,bhattarai2024triplanenet} opens up paths for conducting 3D semantic editing directly in the tri-plane space rather than the GAN parameter space, thereby avoiding the issues mentioned above. In our study, we utilize TriPlaneNet~\cite{bhattarai2024triplanenet} as our Image-to-plane model to transform both the identity image $\mathbf{I}^{id}$ and the reference image $\mathbf{I}^{ref}$ into tri-plane representations, $\mathbf{T}^{id}$ and $\mathbf{T}^{ref}$, respectively, for subsequent use.

\vspace{-0.1in}
\begin{equation}
\label{eq:t_id}
\mathbf{T}^{id} = TriPlaneNet(\mathbf{I}^{id})
\end{equation}
\begin{equation}
\label{eq:t_ref}
\mathbf{T}^{ref} = TriPlaneNet(\mathbf{I}^{ref})
\end{equation}

Having obtained the tri-plane representations, $\mathbf{T}^{id}$ and $\mathbf{T}^{ref}$, our objective is to merge $\mathbf{T}^{ref}$ into $\mathbf{T}^{id}$ by targeting semantic regions that match the semantic category $k$ from the reference images. The method for locating these semantic regions is described in the following paragraph.

\paragraph{Semantic and RGB Decoders with Volume Rendering.}
To generate a semantic map for the reference image, drawing inspiration from pix2pix3D~\cite{deng20233d}, we deploy a semantic decoder $f^{s}$ designed to predict semantics that match the underlying geometry represented in the density volume. The method is outlined as follows: Initially, for each 3D position queried $\mathbf{x}$, we obtain the corresponding feature vector $(\mathbf{f}_{xy}, \mathbf{f}_{xz}, \mathbf{f}_{yz})$ by projecting $\mathbf{x}$ onto each tri-plane of $\mathbf{T}^{ref}$ using bilinear interpolation. These vectors are then summed to form the final features $\mathbf{f}$. Subsequently, the semantic decoder processes the features $\mathbf{f}$ to produce the semantic label $\mathbf{s}$, density $\sigma$, and feature embedding $\mathbf{\phi}$:

\vspace{-0.1in}
\begin{equation}
\label{eq:sampling}
\mathbf{f} = sampling(\mathbf{T}^{ref}, \mathbf{x})
\end{equation}
\begin{equation}
\label{eq:decoder}
(\mathbf{s}, \sigma, \mathbf{\phi})=f^{s}\left(\mathbf{f}\right)
\end{equation}

where $f^{s}$ is a 2-layer MLP, queried $\mathbf{x} \in \mathbb{R}^3$, $K$ the number of semantic category, semantic label $\mathbf{s} \in \mathbb{R}^K$, density $\sigma \in \mathbb{R}^+$, and feature embedding $\mathbf{\phi} \in \mathbb{R}^{64}$. After creating the semantic and density fields, we are able to transform them into a semantic map from any camera angle through classical volume rendering. Specifically, for each pixel, we sample along a camera ray $\mathbf{r}(t) = \mathbf{o} + t\mathbf{d}$, which starts at the camera position $\mathbf{o}$ and extends in the viewing direction $\mathbf{d}$, between the near and far bounds $t_{n}$ and $t_{f}$. Following the approach described by \cite{mildenhall2021nerf,deng20233d}, we sample $N$ points along the ray emanating from a pixel's location. Denote $x_{i}$ as the i-th sampled point on the ray $\mathbf{r}$. Let $\mathbf{s}_{i}$, $\sigma_{i}$, and $\mathbf{\phi}_{i}$ represent the semantic label, density, and feature embedding of $x_{i}$, respectively. We then compute the semantic map $\mathbf{S}^{ref}$ using the volume rendering:
\vspace{-0.1in}
\begin{equation}
\label{eq:t_for_render}
\mathbb{T}_i=\exp \left(-\sum_{j=1}^{i-1} \sigma_j \delta_j\right)
\end{equation}

\vspace{-0.1in}
\begin{equation}
\vspace{-0.1in}
\label{eq:render_s}
\mathbf{S}^{ref}(\mathbf{r})=\sum_{i=1}^N \mathbb{T}_{i} \left(1-\exp\left(-\sigma_{i}\delta_{i}\right) \right) \mathbf{s}_i
\end{equation}


where $\delta_i = t_{i+1} - t_{i}$. At this point, we construct the semantic map for the reference image $\mathbf{I}^{ref}$ with a focus on 3D awareness. It's important to note that our semantic map can be rendered from a different viewpoint, allowing for the alignment of the semantic region from a reference image with a large pose difference into a desired pose. In our case, we render the semantic map $\mathbf{S}^{ref}$ from a frontal perspective to offer details necessary for aligning semantic regions from the reference image $\mathbf{I}^{ref}$ with the identity image $\mathbf{I}^{id}$ in the following paragraph. Additionally, we deploy an RGB decoder $f^{c}$ to reconstruct the RGB image in the frontal view of $\mathbf{I}^{ref}$ following the same processing pipeline used for $f^{s}$. We denote the final reconstructed RGB image as $\hat{\mathbf{I}}^{ref}$, which will be used for appearance transfer in Sec.~\ref{method:coarse_to_fine_image_inpainting}.


\begin{figure*}[t]
    \begin{center}
    \centering
    \includegraphics[width=1.0\linewidth]{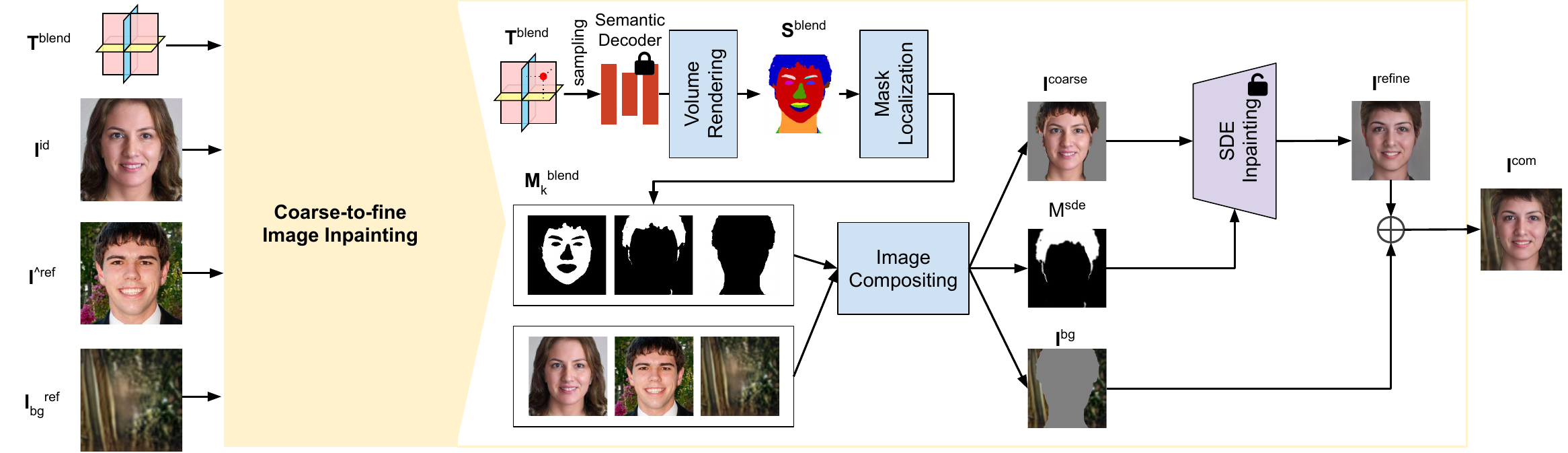}
    \end{center}
\vspace{-0.1in}
\caption{After extracting semantic masks from the blended tri-plane features, we employ a coarse-to-fine inpainting technique to incorporate the reference context into the identity image, ensuring that unmodified areas are preserved to generate the final output.}
\label{fig:pipeline_b}
\end{figure*}

\paragraph{Mask Localization and Tri-plane Blending.}

By utilizing the tri-plane representation $\mathbf{T}^{id}$ and $\mathbf{T}^{ref}$ within the canonical tri-plane space, we not only facilitate 3D semantic editing but also overcome alignment challenges common in 2D image space. This is achieved through the integration of semantic mask localization and tri-plane blending. Specifically, we use a binary mask $\mathbf{M}_{k}^{ref}$ to delineate the local area associated with the desired attribute of semantic category $k$ in $\mathbf{S}^{ref}$. Meanwhile, $\mathbf{M}_{\neg k}^{id}$ define the regions excluding those of semantic category $k$ in $\mathbf{S}^{id}$:





\begin{equation}
\label{eq:mask_ref_k}
\mathbf{M}_{k}^{ref}[x, y]=1\{\mathbf{S}^{ref}[x, y] \in \{k\}\}
\end{equation}
\begin{equation}
\label{eq:mask_id_neq_k}
\mathbf{M}_{\neg k}^{id}[x, y]=1\{\mathbf{S}^{id}[x, y] \notin \{k\}\}
\end{equation}

where $1\{.\}$ is the indicator function. Initially, we remove the exiting attribute of semantic category $k$ in $\mathbf{T}^{id}$ to create $\hat{\mathbf{T}}^{id}$ by applying the mask $\mathbf{M}_{\neg k}^{id}$:

\begin{equation}
\label{eq:t_neq_k}
\mathbf{T}_{\neg k} = \left(\mathbf{M}_{\neg k}^{id}, 1^{XZ}, 1^{YZ} \right)
\end{equation}
\begin{equation}
\label{eq:hat_t_id}
\hat{\mathbf{T}}^{id} = \mathbf{T}^{id} \cdot \mathbf{T}_{\neg k}
\end{equation}

We then transfer the new structure of the attribute from $\mathbf{T}^{ref}$ into $\hat{\mathbf{T}}^{id}$ by projecting $\mathbf{T}^{ref}$ onto the XY-plane, under the guidance of $\mathbf{M}_{k}^{ref}$, and subsequently composite $\mathbf{T}^{blend}$ using alpha blending composition:

\begin{equation}
\label{eq:t_alpha}
\mathbf{T}^{alpha} = \left(\mathbf{M}_{k}^{ref}, 1^{XZ}, 1^{YZ} \right)
\end{equation}
\begin{equation}
\label{eq:t_blend}
\mathbf{T}^{blend} = \hat{\mathbf{T}}^{id} \cdot \left(1 - \mathbf{T}^{alpha}\right) + \mathbf{T}^{ref} \cdot \mathbf{T}^{alpha}
\end{equation}

Upon acquiring the blended tri-plane $\mathbf{T}^{blend}$, we initiate the semantic map $\mathbf{S}^{blend}$ and rendering of novel views for semantic images through designated decoders $f^{s}$, adhering to the methods outlined in Eq.~\ref{eq:sampling}-\ref{eq:render_s}. We illustrate the entire pipeline of Sec.~\ref{method:semantic_guided_triplane_blending} in Fig.~\ref{fig:pipeline_a}.



\subsection{Coarse-to-fine Image Inpainting}
\label{method:coarse_to_fine_image_inpainting}


After transferring the structure associated with the local semantic mask $\mathbf{M}_{k}^{ref}$, we then proceed to transfer the appearance, such as color and texture, from the reconstructed reference image $\hat{\mathbf{I}}^{ref}$ into 2D output image. To achieve this, we propose a coarse-to-fine image inpainting method designed to produce refined images while addressing issues such as color and texture inconsistencies, leakage of facial details, and rough transitions at the boundary between the editing region and the background.

\paragraph{Image Compositing.}

We begin by designing a set of semantic masks, directed by the semantic map $\mathbf{S}^{blend}$, to ensure that the masks outlining the inpainting regions are precisely and smoothly defined. Specifically, our focus is on creating $\mathbf{M}_{k}^{blend}$, $\mathbf{M}_{skin}^{blend}$, and $\mathbf{M}_{bg}^{blend}$, which correspond to the regions for semantic category $k$, skin, and background, respectively. We can then create a coarse composite image $\mathbf{I}^{coarse}$ along with a corresponding background image $\mathbf{I}^{bg}$:



\vspace{-0.1in}
\begin{equation}
\label{eq:i_coarse}
\mathbf{I}^{coarse} = \mathbf{I}^{id} \cdot \mathbf{M}_{\neg bg}^{blend} \cdot \mathbf{M}_{\neg k}^{blend} + \hat{\mathbf{I}}^{ref} \cdot \mathbf{M}_{k}^{blend}
\end{equation}
\begin{equation}
\label{eq:i_bg}
\mathbf{I}^{bg} = \mathbf{I}_{bg}^{ref} \cdot \mathbf{M}_{bg}^{blend}
\end{equation}

where $\mathbf{I}_{bg}^{ref}$ denotes a reference image for the background. However, leakage in the face region may occur because the RGB contexts in $\mathbf{I}^{id}$ and $\hat{\mathbf{I}}^{ref}$ might not completely encompass the target semantic region in $\mathbf{S}^{blend}$. This can be attributed to the varying shapes and scales of the semantic region among $\mathbf{S}^{blend}$, $\mathbf{I}^{id}$, and $\hat{\mathbf{I}}^{ref}$. To address any resulting gaps, we employ a strategy that utilizes patches from existing semantic regions in $\mathbf{I}^{id}$ or $\hat{\mathbf{I}}^{ref}$ to fill in the missing contexts, e.g. skin context filling.


\begin{equation}
\label{eq:m_null}
\mathbf{M}^{null}[x, y] = 1\{\mathbf{I}^{coarse}[x, y] = \mathbf{0}\}
\end{equation}
\begin{equation}
\label{eq:m_gap}
\mathbf{M}^{gap}[x, y] = 1\{\mathbf{M}_{skin}^{blend}[x, y] \cap \mathbf{M}^{null}[x, y]\}
\end{equation}
\begin{equation}
\label{eq:i_coarse_fill_gap}
\mathbf{I}^{coarse}[x, y] = \overline{RGB}_{skin}(\mathbf{I}^{id})\{\mathbf{M}^{gap}[x, y]=1\}
\end{equation}

where $\overline{RGB}_{k}(.)$ retrieves the mean of RGB pixel values corresponding semantic category $k$. At this point, we create a coarse composite image that aligns with the target semantic region $\mathbf{S}^{blend}$ and includes the corresponding RGB context from $\mathbf{I}^{id}$ and $\hat{\mathbf{I}}^{ref}$. In the following paragraph, we will focus on improving context consistency, especially along the borders of the inpainted areas, to minimize abrupt transitions.


\begin{figure*}[t]
    \begin{center}
    \centering
    \includegraphics[width=1.0\linewidth]{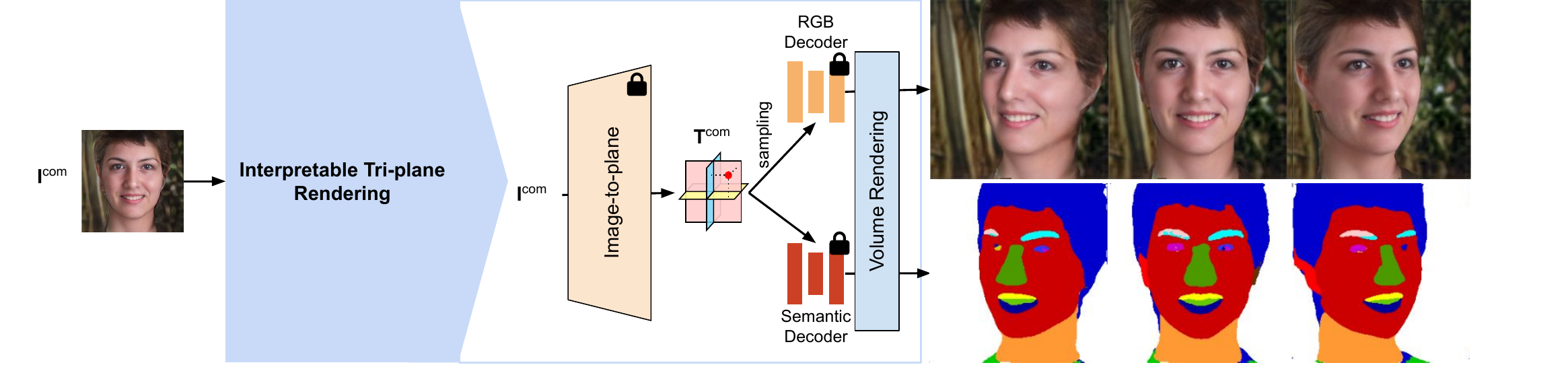}
    \end{center}
\vspace{-0.1in}
\caption{Utilizing the Image-to-plane module, we convert the composite image into a tri-plane representation and subsequently perform volume rendering to generate novel-view RGB and semantic images.}
\label{fig:pipeline_c}
\end{figure*}

\paragraph{SDE Inpainting.}

Drawing inspiration from SDEdit~\cite{meng2021sdedit}, we utilize Stochastic Differential Equations(SDE) as a generative prior along with an iterative denoising process, to facilitate controllable image synthesis and modification. Initially, we create an mask $\mathbf{M}^{sde}$ that augments the semantic mask $\mathbf{M}_{k}^{blend}$ by integrating the gap mask $\mathbf{M}^{gap}$. This is achieved by applying dilation and blurring operations as described in \cite{wu2022hairmapper} to ensure that it covers the boundaries for a smooth transition and clearly defines which areas of the image should be retained or altered. Subsequently, SDEdit introduces Gaussian noise to perturb the coarse image $\mathbf{I}^{coarse}$ and redefines the editing process as the successive removal of this noise. This transition progresses from the disrupted coarse image to the refined image $\mathbf{I}^{refine}$, guided by the reverse SDE that gradually denoises the image. During each iteration, the editing is restricted to regions specified by the mask $\mathbf{M}^{sde}$. The areas being edited are incrementally steered towards a targeted outcome, which draws on the coarse image $\mathbf{I}^{coarse}$ for guidance.

\vspace{-0.1in}
\begin{equation}
\label{eq:m_sde}
\mathbf{M}^{sde} = Blur\left(Dilate\left(\mathbf{M}_{k}^{blend} \cup \mathbf{M}^{gap}\right)\right)
\end{equation}
\begin{equation}
\label{eq:i_refine}
\mathbf{I}^{refine} = SDEdit\left(\mathbf{I}^{coarse}, \mathbf{M}^{sde}, t_{0}, N, K; \theta\right)
\end{equation}

where $t_{0}$ serves as a scalar within the SDE that sets the level of noise and is a value within the range (0, 1]. $N$ represents the total number of steps in the denoising process. $K$ indicates the total number of iterations the entire process is run. $\theta$ denotes the parameters used in the score-based diffusion model. After refining the image to obtain $\mathbf{I}^{refine}$, we blend it with $\mathbf{I}^{bg}$ to create the composite image $\mathbf{I}^{com}$.

\begin{equation}
\label{eq:i_com}
\mathbf{I}^{com} =  \mathbf{I}^{refine} \cdot \mathbf{M}_{\neg bg}^{blend} + \mathbf{I}^{bg}
\end{equation}

The complete workflow of Sec.~\ref{method:coarse_to_fine_image_inpainting} is depicted in Fig.~\ref{fig:pipeline_b}.

\subsection{Interpretable Tri-plane Rendering}
\label{method:interpretable_triplane_rendering}

We now have $\mathbf{I}^{com}$ with both structure and appearance correctly aligned in the frontal view. Finally, We convert $\mathbf{I}^{com}$ into tri-plane representation $\mathbf{T}^{com}$ using \textbf{Image-to-plane} method in Sec.\ref{method:semantic_guided_triplane_blending}. This subsequently enables the rendering of novel views for RGB and semantic images, utilizing the \textbf{Semantic and RGB Decoders with Volume Rendering} processes detailed in Sec.\ref{method:semantic_guided_triplane_blending}. The full sequence of procedures is presented visually in Fig.~\ref{fig:pipeline_c}.

\subsection{Learning Objectives}
\label{method:training_objectives}

We utilize stage training, dividing the training process into two phases to concentrate on varying aspects of detail within the images, thereby progressively enhancing the performance of edited image synthesis. In the first phase, we train the RGB decoder $f^{c}$ and semantic decoder $f^{s}$ focusing on the objectives of \textbf{Reconstruction Loss} and \textbf{Adversarial Loss}. In the second phase, we train the score-based diffusion model, parametrized by $\theta$, with the objective of \textbf{Denoising diffusion Objective}.



\paragraph{Reconstruction Loss.}
The purpose of reconstruction loss is to measure the difference between the ground-truth and their reconstructed counterparts. Given a camera ray $\mathbf{r}$ from a ground-truth viewpoint associated with the original RGB image $\mathbf{I}_{\mathbf{c}}$ and semantic map $\mathbf{I}_{\mathbf{s}}$, we render the reconstructed RGB image $\hat{\mathbf{I}}^{ref}$ and semantic map $\mathbf{S}^{ref}$ and then calculate reconstruction losses. Following pix2pix3D~\cite{deng20233d}, we use LPIPS~\cite{zhang2018unreasonable} to compute the image reconstruction loss $\mathcal{L}_{c}$ for RGB images $\hat{\mathbf{I}}^{ref}$. For semantic label reconstruction loss $\mathcal{L}_{s}$, we use the balanced cross-entropy loss for segmentation maps $\mathbf{S}^{ref}$.

\vspace{-0.1in}
\begin{equation}
\label{eq:l_c}
\mathcal{L}_{c} = LPIPS\left(\mathbf{I}_{\mathbf{c}}, \hat{\mathbf{I}}^{ref} \right)
\end{equation}

\vspace{-0.1in}
\begin{equation}
\label{eq:l_s}
\mathcal{L}_{s} =-\sum_{i=1}^N \sum_{k=1}^K w_{k} \cdot y_{i, k} \log \left(\hat{y}_{i, k}\right)
\end{equation}

\vspace{-0.1in}
\begin{equation}
\label{eq:l_recon}
\mathcal{L}_{\text {recon }}=\lambda_{c} \mathcal{L}_{c}+\lambda_{s} \mathcal{L}_{s}
\end{equation}

where $N$ represents the total number of pixels in the semantic maps $\mathbf{S}^{ref}$. $K$ denotes the number of semantic labels. $y_{i, k}$ is a binary indicator (0 or 1) indicating whether semantic category $k$ is the correct classification for pixel $i$. $\hat{y}_{i, k}$ is the predicted probability that pixel $i$ is classified as semantic category $k$. $w_{k}$ is the weight assigned to semantic category $k$. $\lambda_{c}$ and $\lambda_{s}$ are hyperparameters that weight the loss terms $\mathcal{L}_{c}$ and $\mathcal{L}_{s}$, respectively.

\paragraph{Adversarial Loss.}
This type of loss is designed to enable Generative Adversarial Networks (GANs)~\cite{goodfellow2020generative} to produce new images that are indistinguishable from real images, thereby improving the realism and quality of the generated images. We employ adversarial loss~\cite{goodfellow2014generative}, denoted as $\mathcal{L}_{GAN}$, to make the discriminator maximize the probability of correctly classifying the input images, while simultaneously encouraging the generator to maximize the error of the discriminator in distinguishing fake images as real. Additionally, following  pix2pix3D~\cite{deng20233d}, we implement an additional discriminator $D_{\mathbf{s}}$ that processes both RGB images and semantic maps. This is designed to promote pixel alignment between RGB images and semantic maps, establishing an auxiliary adversarial loss $\mathcal{L}_{aux}$.


\begin{equation}
\label{eq:i_com_hat_i_com}
\mathbf{I}_{con} = [\mathbf{I}_{\mathbf{c}}, \mathbf{I}_{\mathbf{s}}], \hat{\mathbf{I}}_{con}= [\hat{\mathbf{I}}^{ref},\mathbf{S}^{ref}]
\end{equation}





\vspace{-0.15in}
\begin{equation}
\label{eq:l_gan}
\mathcal{L}_{GAN} = \mathbb{E}_{\mathbf{I}_{c}}[\log D_{\mathbf{c}}(\mathbf{I}_{c})] + \mathbb{E}_{\hat{\mathbf{I}}^{ref}}[\log (1-D_{\mathbf{c}}(\hat{\mathbf{I}}^{ref}))]
\end{equation}

\vspace{-0.2in}
\begin{equation}
\label{eq:l_aux}
\mathcal{L}_{aux} = \mathbb{E}_{\mathbf{I}_{con}}[\log D_{\mathbf{s}}(\mathbf{I}_{con})] + \mathbb{E}_{\hat{\mathbf{I}}_{con}}[\log (1-D_{\mathbf{s}}(\hat{\mathbf{I}}_{con}))]
\end{equation}

\begin{equation}
\label{eq:adv_loss}
\min_{f^{c}, f^{s}}{\max_{D_{\mathbf{c}}, D_{\mathbf{s}}} \lambda_{D_{\mathbf{c}}}{\mathcal{L}_{GAN}} + \lambda_{D_{\mathbf{s}}}\mathcal{L}_{aux}}
\end{equation}


where [.] denotes a concatenation operation. $\lambda_{D_{\mathbf{c}}}$ and $\lambda_{D_{\mathbf{s}}}$ are hyperparameters used to weight the loss terms $ \mathcal{L}_{GAN}$ and $\mathcal{L}_{aux}$, respectively.

\paragraph{Denoising diffusion Objective}



The \textbf{SDE Inpainting} process outlined in Sec.\ref{method:coarse_to_fine_image_inpainting} utilizes a reverse SDE, detailed in \cite{meng2021sdedit,song2020score}, which transitions from $t = 1$ to $t = 0$ to progressively remove noise and yield a denoised result that is both realistic and accurate. This process is informed by the noise-perturbed score function, which is estimated using a parametrized score model developed through denoising score matching~\cite{vincent2011connection}. We refer to this parametrized score model as $\boldsymbol{s}_{\boldsymbol{\theta}}\left(\mathbf{x}\left(t\right), t\right)$. The overall training objective is defined as:

\vspace{-0.1in}
\begin{equation}
\label{eq:x_t}
\mathbf{x}(t)=\alpha(t) \mathbf{x}+\sigma(t) \mathbf{z}
\end{equation}

\vspace{-0.1in}
\begin{equation}
\label{eq:l_sde}
L_{SDE}(\mathbf{x}, \theta)=\mathbb{E}_{t, \mathbf{z} \sim \mathcal{N}(\mathbf{0}, \mathbf{I})}\left[\left\|\boldsymbol{s}_{\boldsymbol{\theta}}(\mathbf{x}(t), t)-\mathbf{z}\right\|_{2}^{2}\right]
\end{equation}

where $\mathbf{x}(t)$ is the version of the input $\mathbf{x}$ that has been perturbed with noise, obtained by sampling $\mathbf{z}$ from a normal distribution $\mathcal{N}(\mathbf{0}, \boldsymbol{I})$, and $t$ is uniformly sampled from the interval $0$ to $1$. $\alpha(t)$ is a scalar function that denotes the magnitude of the input $\mathbf{x}$, while $\sigma(t)$ is a scalar function that describes the magnitude of the noise $\mathbf{z}$.

\section{Experiments}
\label{sec:experiments}

In this section, we first present both quantitative and qualitative evaluations compared with competing latent-based and reference-based image editing methods in Sec.~\ref{exp:comparisons}. Subsequently, we demonstrate various applications utilizing our edited triplane representation, guided by reference images, in Sec.~\ref{exp:application}.

\paragraph{Datasets.}

We use FFHQ~\cite{karras2019style} as our training dataset, which contains 70,000 facial images featureing a wide variety of human faces, covering different ages, ethnicities, and image backgrounds. Following~\cite{zhu2021barbershop}, we utilize a pre-trained face parsing model~\cite{yu2018bisenet}, which was trained using the CelebAMask-HQ~\cite{lee2020maskgan}, to generate semantic maps for the FFHQ dataset. These semantic maps categorize 19 classes, such as nose, eyebrows, ears, eyeglasses, lips, etc. The CelebA~\cite{liu2015faceattributes} dataset consists of 202,599 images, each annotated with 40 distinct attributes. These annotations cover a wide range of features, including hair type, eyeglasses, facial expressions, and more. For our experiments, we utilize CelebA to conduct both quantitative and qualitative evaluations in Sec.~\ref{exp:comparisons}. We resize all images to the size of 256×256 for our experiments.

\paragraph{Implementation details.}

For the Image-to-plane model, we utilize a pre-trained TriPlaneNet~\cite{bhattarai2024triplanenet}. In terms of RGB and semantic rendering, we employ 2-layer Multi-Layer Perceptrons (MLP) for $f^{s}$ and $f^{c}$. The discriminators $D_{c}$ and $D_{s}$ are implemented following the architecture detailed in~\cite{chan2022efficient}. We set the values for $\lambda_{c}$, $\lambda_{s}$, $\lambda_{D_{\mathbf{c}}}$, and $\lambda_{D_{\mathbf{s}}}$ to $1$. For SDE inpainting, we train a score-based diffusion model using the Variance Preserving (VP) SDE architecture described in~\cite{meng2021sdedit}, ensuring that $\alpha^{2}(t)+\sigma^{2}(t)=1$ for all $t$, with $\alpha(t)$ approaching $0$ as $t$ approaches $1$. The model training is conducted using 4 NVIDIA A100 GPUs.

\begin{figure*}[t]
    \begin{center}
    \centering
        \includegraphics[width=0.6\linewidth]{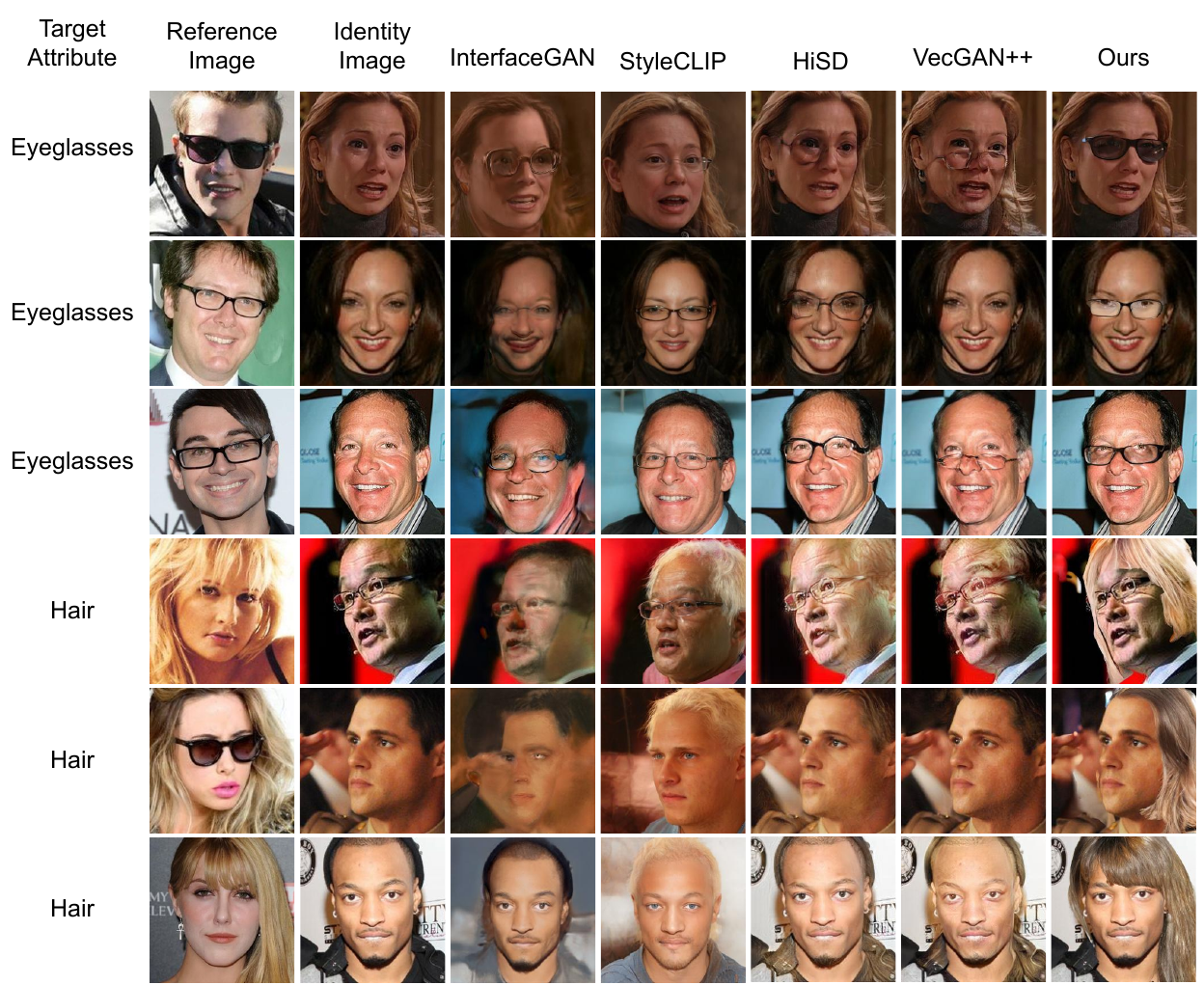}
    \end{center}
\vspace{-0.1in}
\caption{Qualitative comparison with other competing editing methods. The top three rows display the addition of eyeglasses, while the remaining rows illustrate changes to hair.}
\label{fig:2d_qualitative_evaluation}
\end{figure*}

\subsection{Comparisons}
\label{exp:comparisons}

\paragraph{Baselines.}

We benchmark our editing method against reference-based image editing methods such as HiSD~\cite{li2021image}, VecGAN++~\cite{dalva2022vecgan}, and Bilecen \textit{et al.}~\cite{bilecen2024reference}, as well as latent-based editing methods including InterfaceGAN~\cite{shen2020interpreting} and StyleCLIP~\cite{patashnik2021styleclip}.

\paragraph{Quantitative evaluation.}

\begin{table*}[]
\centering
\begin{tabular}{lllllll}
\hline
\multicolumn{1}{c}{\multirow{2}{*}{Method}} & \multicolumn{3}{c}{Eyeglasses} & \multicolumn{3}{c}{Hair} \\ \cline{2-7} 
\multicolumn{1}{c}{}                        & FID $\downarrow$     & SSIM $\uparrow$    & $\mathcal{L}_{2} \downarrow$      & FID $\downarrow$    & SSIM $\uparrow$    & $\mathcal{L}_{2} \downarrow$    \\ \hline
InterfaceGAN~\cite{shen2020interpreting} & 88.13 & 0.9398 & 0.0104 & 80.93 & 0.7888 & 0.0387 \\
StyleCLIP~\cite{patashnik2021styleclip} & 80.13 & 0.8421 & 0.0476 & 92.60 & 0.8716 & 0.0196 \\ \hline
HiSD~\cite{li2021image} & 77.56 & 0.9471 & 0.0090 & 94.53 & 0.9743 & 0.0036 \\
VecGAN++~\cite{dalva2022vecgan} & 71.47 & 0.7483 & 0.0630 & 80.47 & 0.9296 & 0.0090 \\
Bilecen \textit{et al.}~\cite{bilecen2024reference} & 66.68 & 0.9818 & \textbf{0.0021} & 64.59 & 0.9720 & \textbf{0.0029} \\

Ours & \textbf{37.87} & \textbf{0.9881} & 0.0049 & \textbf{42.74} & \textbf{0.9932} & 0.0038 \\
\hline
\end{tabular}
\caption{Quantitative evaluation on two image editing tasks: the addition of eyeglasses and altering hair color from black to blonde.}
\label{exp:quantitative_evaluation}
\end{table*}

Following the evaluation protocol outlined in Bilecen \textit{et al.}~\cite{bilecen2024reference}, we evaluate our editing method through two distinct tasks: adding eyeglasses and changing hair color from black to blonde. For the eyeglasses addition task, we categorize our test image sets into two groups: those with eyeglasses and those without. We then introduce eyeglasses to the images lacking them, and subsequently, we compute the Frechet Inception Distances (FIDs)~\cite{heusel2017gans} between these edited images and the original images that already have eyeglasses. We apply the same process for the hair color editing task as well. As shown in Tab.~\ref{exp:quantitative_evaluation}, our method significantly surpasses competing approaches in terms of FID across both tasks. Besides evaluating quality with FID, we utilize the Structural Similarity Index Measure (SSIM) and normalized L2 distance metrics to assess reconstruction quality after editing. Adhering to the evaluation protocol in Bilecen \textit{et al.}~\cite{bilecen2024reference}, we exclude the areas modified during editing, which are identified as eyeglasses or hair by the segmentation network, and measure the pixel differences between the input and edited images. As demonstrated in Tab.~\ref{exp:quantitative_evaluation}, our method achieves the highest performance in terms of SSIM compared to other competing approaches. Although our L2 distance results shows a slight degradation when compared to the results from Bilecen \textit{et al.}~\cite{bilecen2024reference}, we prioritize SSIM because it captures more structural information and is likely to align better with perceived visual quality than the L2 distance metric.



\paragraph{Qualitative evaluation.}


As illustrated in Fig.~\ref{fig:2d_qualitative_evaluation}, latent-based editing methods such as InterfaceGAN~\cite{shen2020interpreting} and StyleCLIP~\cite{patashnik2021styleclip} are capable of adding eyeglasses and changing hair color by manipulating latent code directions within StyleGAN's W+ space. However, these methods fall short in reference-based image editing tasks, such as precisely matching the appearance of eyeglasses to those in a reference image. On the other hand, reference-based editing methods like HiSD~\cite{li2021image} and VecGAN++~\cite{dalva2022vecgan} face challenges in maintaining fidelity to the eyeglasses shown in reference images, particularly when there are significant variations in facial pose between the source and reference images. This issue also arises in tasks involving hair changes. Our method successfully closes the gap between these two types of approaches, seamlessly adding eyeglasses from the reference image to the source image and transferring the hairstyle from the reference image with promising fidelity. Notably, we do not conduct a qualitative comparison with Bilecen \textit{et al.}~\cite{bilecen2024reference} due to the unavailability of their model and the insufficient details provided for reproduction.





\paragraph{Abalation Study and Additional Visualization Results.}

We conduct an ablation study to evaluate the impact of individual modules in our proposed technique and provide additional visualization results. Please refer to Sec. A and B of the Supplemental Material for further details.

\subsection{Application}
\label{exp:application}

 In this section, we employ Next3D~\cite{sun2023next3d}, a 3D-aware facial avatar generator, to generate the identity image and modify it using reference images from FFHQ to demonstrate various applications of our editing method.



\paragraph{Semantic-aware Facial attribute Editing.}

In addition to the qualitative outcomes of eyeglasses and hair editing presented in Sec.~\ref{exp:comparisons}, we showcase a total of seven types of facial attribute edits with novel view RGB and semantic renderings as displayed in Fig.~\ref{fig:semantic_editing}. Importantly, our method accurately aligns semantic regions from the reference images with the identity image, even when the poses of the reference images vary. Moreover, we highlight additional applications in Sec. C of the Supplemental Material.

\section{Conclusion}
\label{sec:conclusion}
In summary, our framework presents a robust solution for local facial attribute editing, overcoming challenges in latent-based and reference-based editing while maintaining 3D consistency and natural appearances. By integrating tri-plane representation and semantic masks, our approach ensures precise alignment and context preservation throughout the editing process. Additionally, our coarse-to-fine inpainting pipeline enhances context consistency and reduces artifacts. Through comprehensive evaluations, we have demonstrated the effectiveness of our method in achieving accurate and realistic results compared to previous methods.


{\small
\bibliographystyle{ieee}
\bibliography{egbib}
}

\clearpage

\setcounter{figure}{5}
\setcounter{table}{1}

\appendix
\thispagestyle{empty}
\twocolumn[{
    \newpage
    \null
    \vskip .375in
    \begin{center}
          {\Large \bf A Reference-Based 3D Semantic-Aware Framework for Accurate Local Facial Attribute Editing \\ Supplemental Material \par}
    
          \vspace*{24pt}
          {
          \large
          \lineskip .5em
          \begin{tabular}[t]{c}
            Yu-Kai Huang, Yutong Zheng, Yen-Shuo Su, Anudeepsekhar Bolimera, Han Zhang, Fangyi Chen \\
            and Marios Savvides\\
            Carnegie Mellon University\\
            5000 Forbes Ave, Pittsburgh, PA 15213\\
            \tt\small yukaih2@andrew.cmu.edu, yutongzh@alumni.cmu.edu, ericsuece@cmu.edu,\\
            \tt\small\{abolimer,hanz3,fangyic,marioss\}@andrew.cmu.edu
          \end{tabular}
          \par
          }
          \vskip .5em
          \vspace*{12pt}
    \end{center}
}]

\section{Abalation Study}
\label{common:ablation}

We perform an ablation study to analyze the effect of individual modules in our proposed technique, focusing on the Coarse-to-fine Image Inpainting stage and the SDE Inpainting module. By comparing the performance of our framework with and without the SDE Inpainting module, we quantitatively assess its contribution to the final output. The results, presented in Tab.~\ref{exp:abalation_study}, show a significant improvement in FID quality when the SDE Inpainting module is included, confirming its importance in enhancing overall performance. Notably, we do not use SSIM and L2 distance metrics because the SDE Inpainting module contributes to the fidelity of the target semantic edited region rather than the reconstruction of unchanged semantic regions.

\begin{table}[h]
\centering
\begin{tabular}{lll}
\hline
\multicolumn{1}{c}{\multirow{2}{*}{Method}} & \multicolumn{1}{c}{Eyeglasses} & \multicolumn{1}{c}{Hair} \\ \cline{2-3} 
\multicolumn{1}{c}{}                        & FID $\downarrow$               & FID $\downarrow$ \\ \hline
w/o SDE & 42.40 & 87.78 \\
w SDE & \textbf{37.87} & \textbf{42.74} \\
\hline
\end{tabular}
\caption{Ablation study of incorporating the SDE Inpainting module in the Coarse-to-fine Image Inpainting stage.}
\label{exp:abalation_study}
\end{table}

\section{Additional Visualization Results of Local Facial Attribute Editing Tasks}
\label{common:additional_visualization}
\begin{figure*}[t]
  \centering
  \begin{subfigure}{0.49\linewidth}
    \includegraphics[width=1.0\textwidth,height=1.0\textwidth]{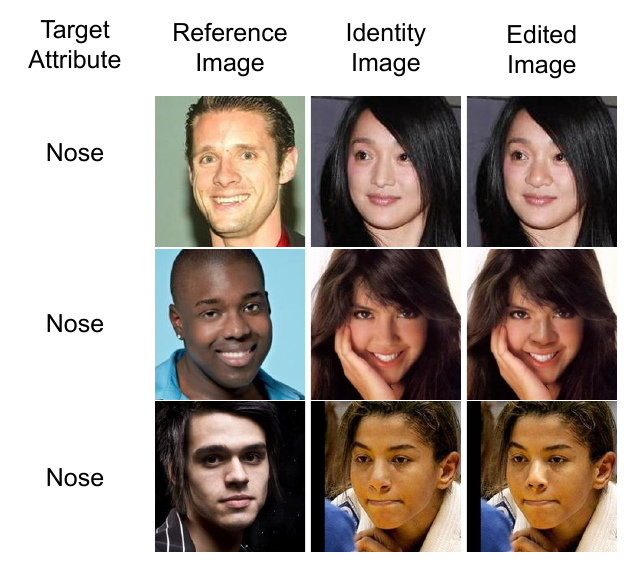}
    \caption{changing the nose to a big nose.}
    \label{fig:nose_editing_task_visualization}
  \end{subfigure}
  \begin{subfigure}{0.49\linewidth}
    \includegraphics[width=1.0\textwidth,height=1.0\textwidth]{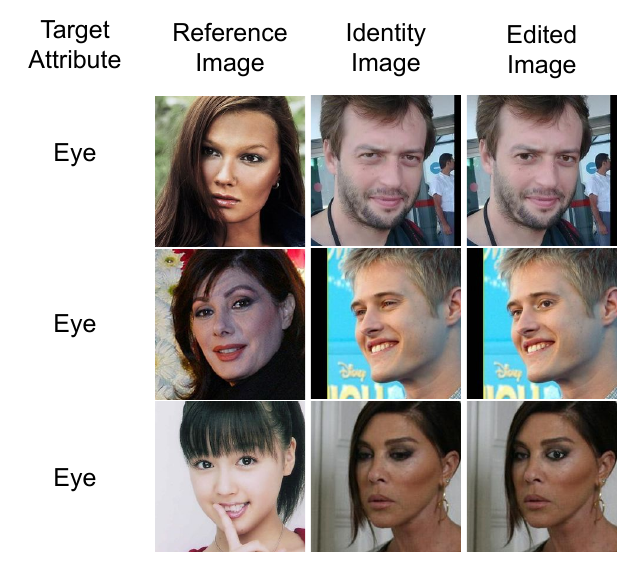}
    \caption{changing the eyes to wide open eyes.}
    \label{fig:eye_editing_task_visualization}
  \end{subfigure}
  \begin{subfigure}{0.49\linewidth}
    \includegraphics[width=1.0\textwidth,height=1.0\textwidth]{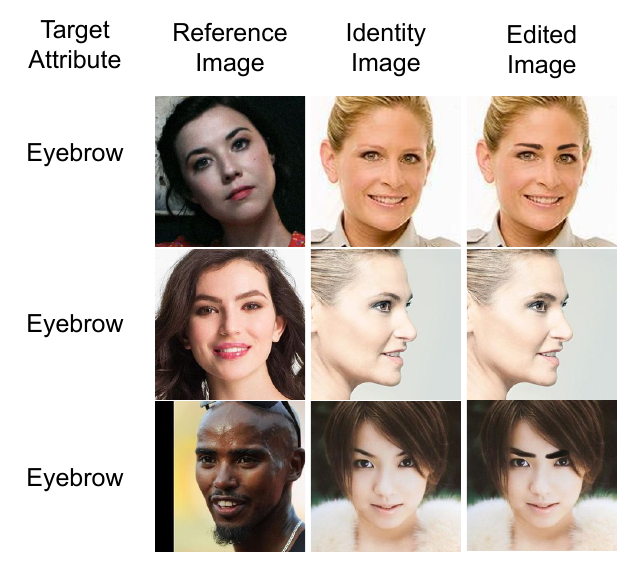}
    \caption{changing the eyebrows to bushy eyebrows.}
    \label{fig:eyebrow_editing_task_visualization}
  \end{subfigure}
  \begin{subfigure}{0.49\linewidth}
    \includegraphics[width=1.0\textwidth,height=1.0\textwidth]{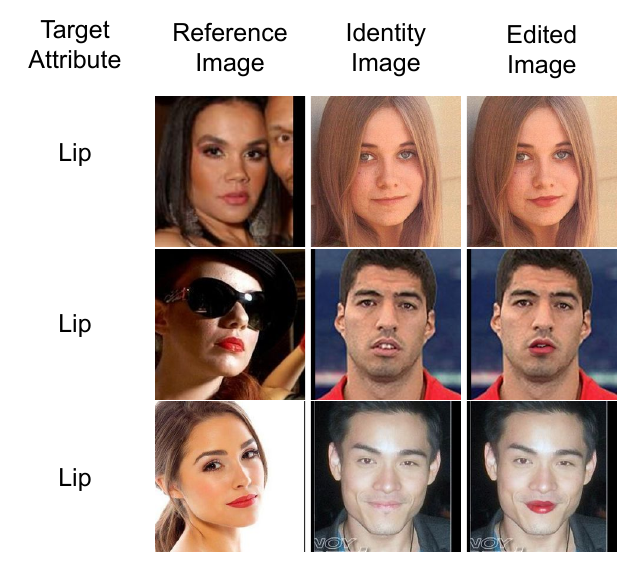}
    \caption{changing the lips to big lips.}
    \label{fig:lip_editing_task_visualization}
  \end{subfigure}
  \caption{Additional Visualization Results of Local
Facial Attribute Editing Tasks.}

  \vspace{-0.5cm}
  \label{fig:exp_other_editing_task_visualization}
\end{figure*}

Regarding the evaluation of local facial attribute editing tasks, there are currently no established quantitative and qualitative benchmarks for attributes other than eyeglasses and hair, to our best knowledge. Consequently, we focus on these two tasks in our evaluations with baseline methods. However, to demonstrate our technique's effectiveness, we also provide additional visualization results for other editing tasks, including changing the nose to a big nose, eyes to wide open eyes, eyebrows to bushy eyebrows, and lips to big lips, as shown in Fig.~\ref{fig:exp_other_editing_task_visualization}.

\begin{figure}[t]
    \begin{center}
    \centering
        \includegraphics[width=\columnwidth]{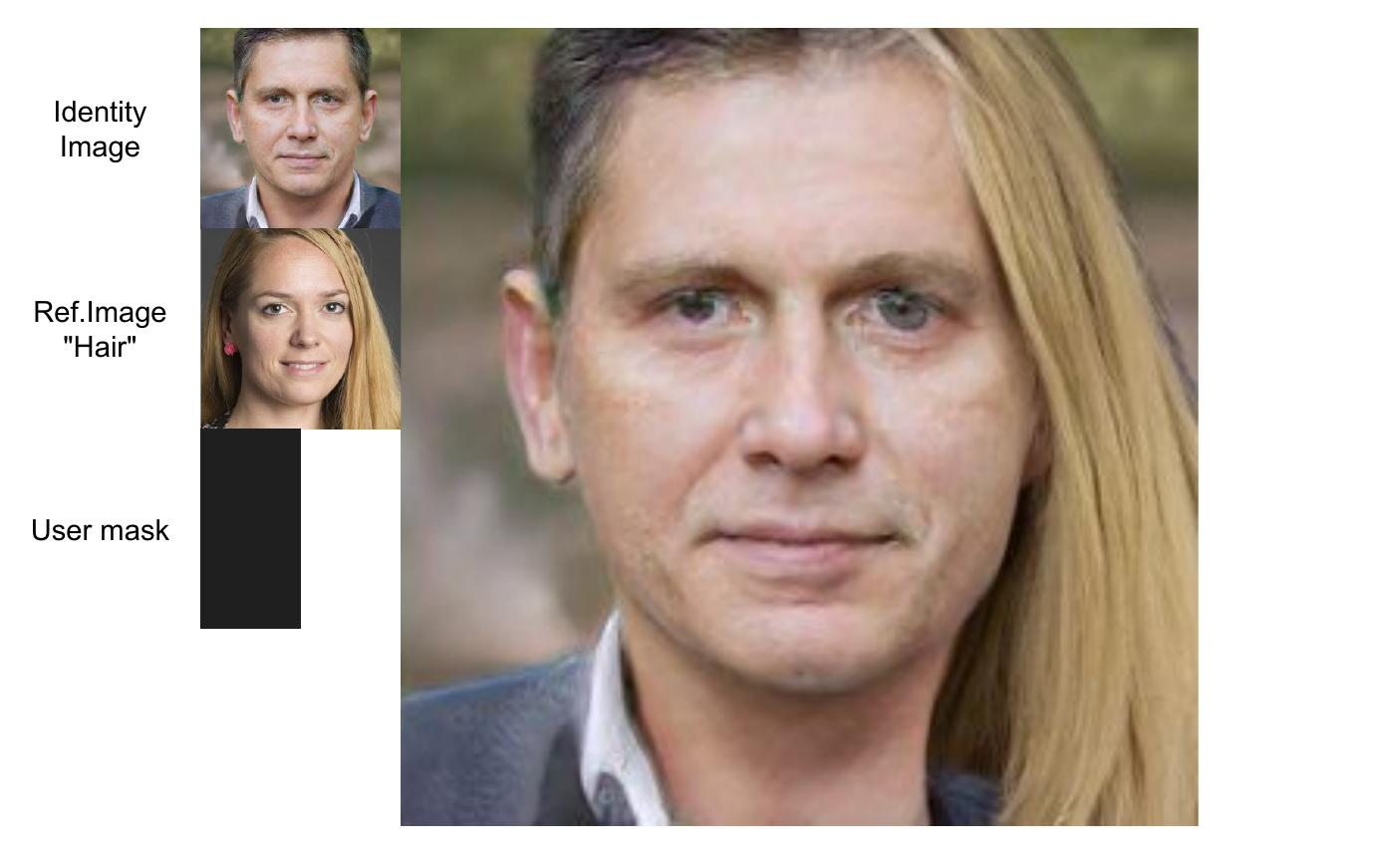}
    \end{center}
\caption{Our editing method allows users to input their own masks for further refinement of target masks generated from predicted semantic maps, enabling precise editing like modifying only half of the hair.}
\label{fig:user_input_mask}
\end{figure}

\section{Additional applications}
\label{exp:additional_application}

\paragraph{Novel view Rendering.}

\begin{figure*}[t]
    \begin{center}
    \centering
        \includegraphics[width=1\linewidth]{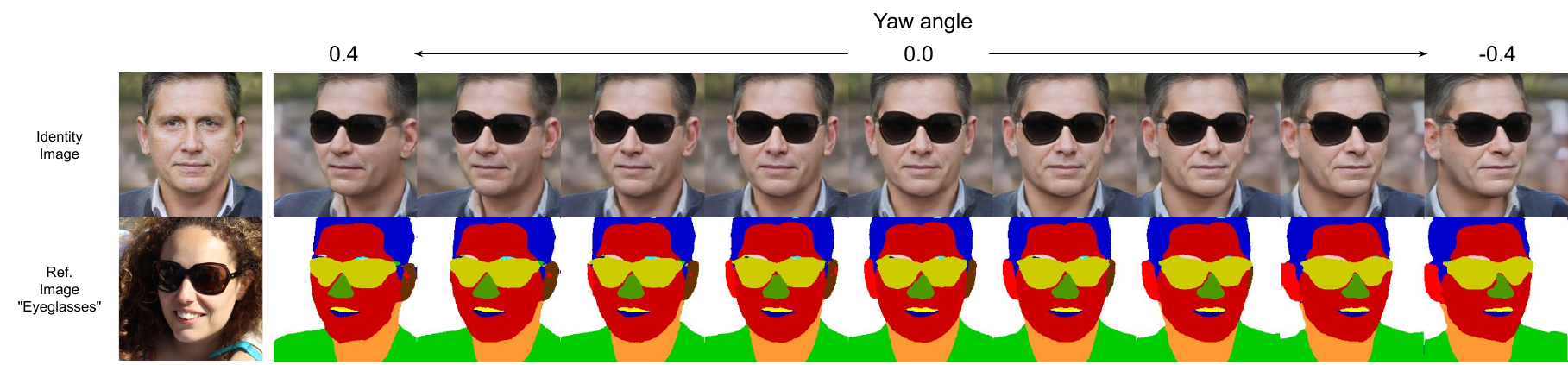}
    \end{center}
\caption{We demonstrate that our method effectively produces continuous RGB and semantic renderings across various poses following editing, modeling both the edited context and semantics within a 3D volume.}
\label{fig:novel_view_rendering}
\end{figure*}

As illustrated in Fig.~\ref{fig:novel_view_rendering}, we effectively produce continuous RGB and semantic renderings across various poses following editing, demonstrating that both the edited context and semantics are modeled within a 3D volume.

\paragraph{Multi-reference Editing.}

\begin{figure}[t]
    \begin{center}
    \centering
        \includegraphics[width=0.8\columnwidth]{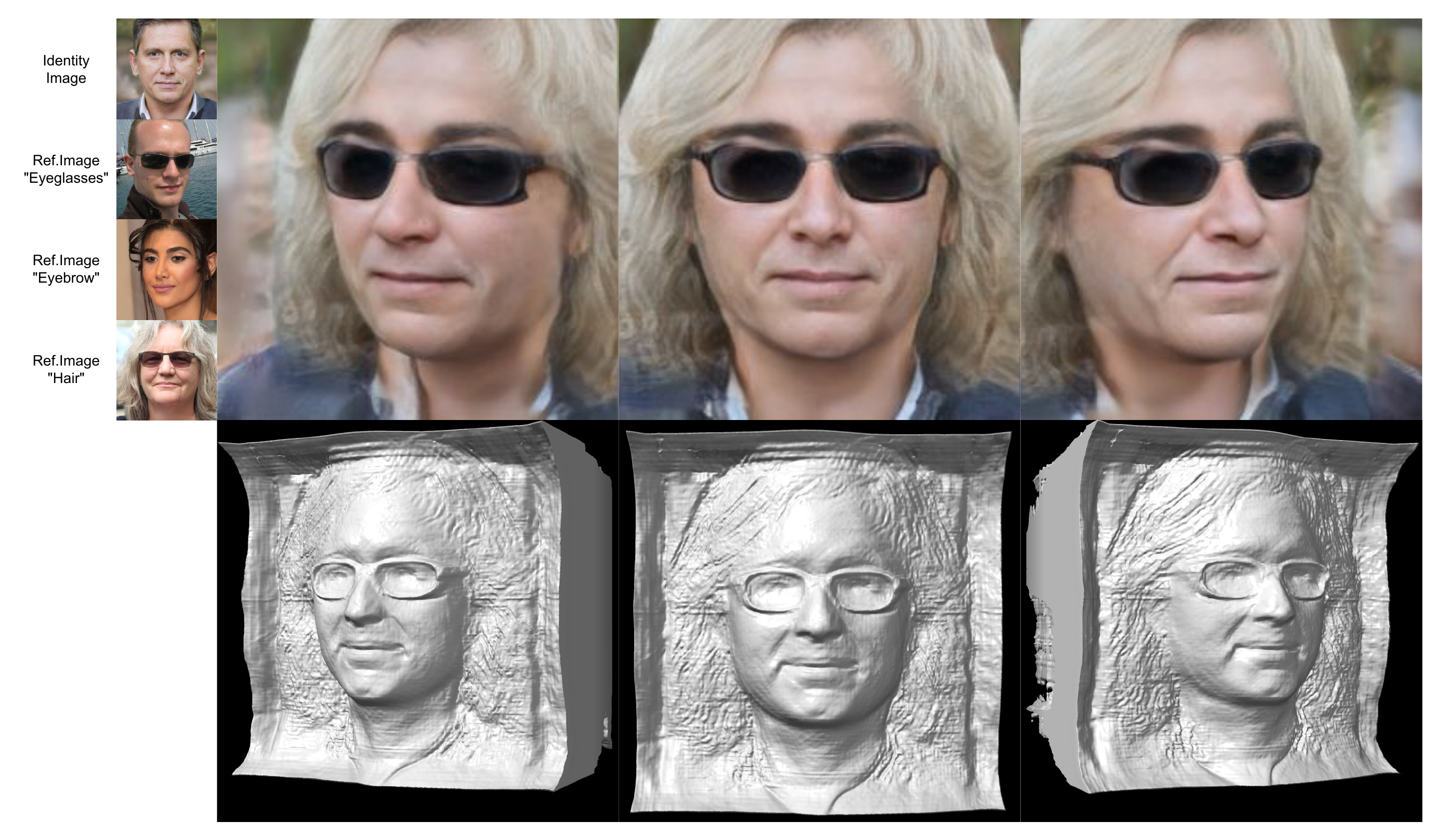}
    \end{center}
\caption{Our method seamlessly combines multiple editing tasks from different reference images simultaneously, maintaining complex geometries and high detail in the resulting shapes and structures.}
\label{fig:multi_reference_editing_w_shape}
\end{figure}

As demonstrated in Fig.~\ref{fig:multi_reference_editing_w_shape}, our method is capable of seamlessly combining multiple editing tasks from different reference images simultaneously, rather than displaying just one facial attribute edit at a time.

\paragraph{Interactive input mask Editing}

As displayed in Fig.~\ref{fig:user_input_mask}, in addition to utilizing target masks directly generated from predicted semantic maps, our editing method also enables users to input their own masks. This allows for further refinement of the target mask and accurately represents the final desired editing result, such as editing only half of the hair.

\paragraph{Geometry Generation.}

As illustrated in Fig.~\ref{fig:multi_reference_editing_w_shape}, the shape and structure resulting from our geometry generation process demonstrate that our editing method is capable of maintaining complex geometries and a high degree of detail.

\paragraph{Reenactment.}

\begin{figure}[t]
    \begin{center}
    \centering
        \includegraphics[width=0.8\columnwidth]{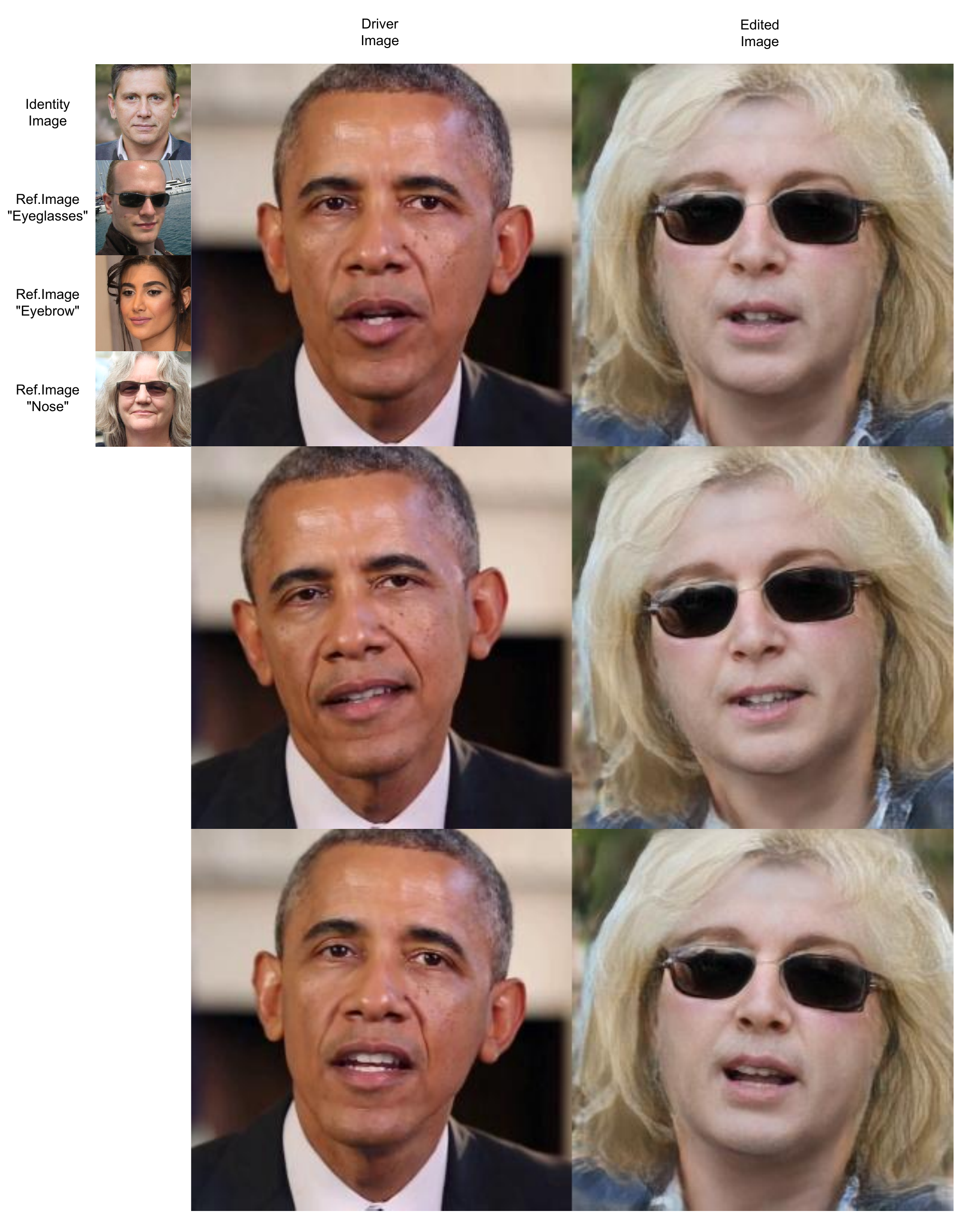}
    \end{center}
\caption{Our editing method, leveraging the FLAME template and deformation parameters for eye and jaw poses and expressions, collaborates with Next3D to enable precise local facial attribute editing with animatable facial generation.}
\label{fig:reenactment}
\end{figure}

Leveraging the FLAME template to construct a foundational mesh that is manipulated by deformation parameters, including eye and jaw poses as well as expressions, our editing method works in conjunction with Next3D [32]. This collaboration facilitates facial attribute editing with animatable facial generation, enabling precise control over expressions and poses simultaneously. This capability is demonstrated in Fig.~\ref{fig:reenactment}.

\section{Future Work}
\label{common:future_work}

In our future work, we plan to integrate image editing with data augmentation for face recognition, adjusting facial attributes while preserving identity to enhance system robustness and accuracy. Additionally, we aim to improve detection techniques for Face Morphing Attacks (FMAs) by creating realistic morphed images from synthetic samples. These samples will train and test detection algorithms, improving the ability to identify and mitigate morphing attacks. We believe these directions will significantly enhance the practical relevance and impact of our work in real-world biometric applications.

\end{document}